\documentclass[letterpaper]{article} 
\usepackage{aaai2026-preprint}
\usepackage{times}  
\usepackage{helvet}  
\usepackage{courier}  
\usepackage[hyphens]{url}  
\usepackage{graphicx} 
\usepackage{natbib}  
\usepackage{caption} 
\usepackage{booktabs}      
\usepackage{xcolor}        
\usepackage{array} 
\usepackage{algorithm}
\usepackage{algorithmic}
\usepackage[utf8]{inputenc}
\usepackage{url}
\usepackage{booktabs}
\usepackage{amssymb}
\usepackage{bbding}
\usepackage{pifont}
\usepackage{utfsym}
\usepackage{newfloat}
\usepackage{listings}
\DeclareCaptionStyle{ruled}{labelfont=normalfont,labelsep=colon,strut=off} 
\floatstyle{ruled}
\newfloat{listing}{tb}{lst}{}
\floatname{listing}{Listing}
\usepackage{amsmath}
\usepackage{algorithm}
\usepackage{algorithmic}
\usepackage{amssymb}
\usepackage{mathrsfs}
\usepackage{bm}

\usepackage{amsthm}

\newtheorem{theorem}{Theorem}

\title{C3D-AD: Toward Continual 3D Anomaly Detection via Kernel Attention with Learnable Advisor}

\author{
    Haoquan Lu\textsuperscript{\rm 1}, Hanzhe Liang\textsuperscript{\rm 2,3}, Jie Zhang\textsuperscript{\rm 4}, Chenxi Hu\textsuperscript{\rm 2}, Jinbao Wang\textsuperscript{\rm 5,6}, Can Gao\textsuperscript{\rm 1,6} \\
}
\affiliations{
    \textsuperscript{\rm 1}College of Computer Science and Software Engineering, Shenzhen University\\
    \textsuperscript{\rm 2}Shenzhen Audencia Financial Technology Institute, Shenzhen University\\
    \textsuperscript{\rm 3}Ningbo Institute of Digital Twin, Eastern Institute of Technology\\
    \textsuperscript{\rm 4}Faculty of Applied Sciences, Macao Polytechnic University\\
    \textsuperscript{\rm 5}School of Artificial Intelligence, Shenzhen University\\
    \textsuperscript{\rm 6}Guangdong Provincial Key Laboratory of Intelligent Information Processing
}

\usepackage{bibentry}

\begin{document}

\maketitle

\begin{abstract}
3D Anomaly Detection (AD) has shown great potential in detecting anomalies or defects of high-precision industrial products. However, existing methods are typically trained in a class-specific manner and also lack the capability of learning from emerging classes. 
In this study, we proposed a continual learning framework named \textbf{C}ontinual \textbf{3D} \textbf{A}nomaly \textbf{D}etection (C3D-AD), which can not only learn generalized representations for multi-class point clouds but also handle new classes emerging over time.
Specifically, in the feature extraction module, to extract generalized local features from diverse product types of different tasks efficiently, Kernel Attention with random feature Layer (KAL) is introduced, which normalizes the feature space. Then, to reconstruct data correctly and continually, an efficient Kernel Attention with learnable Advisor (KAA) mechanism is proposed, which learns the information from new categories while discarding redundant old information within both the encoder and decoder. Finally, to keep the representation consistency over tasks, a Reconstruction with Parameter Perturbation (RPP) module is proposed by designing a representation rehearsal loss function, which ensures that the model remembers previous category information and returns category-adaptive representation.
The proposed method is the first attempt to address 3D anomaly detection in a class-incremental manner, providing the capabilities of multi-class and continual anomaly detection.
Extensive experiments on three public datasets demonstrate the effectiveness of the proposed method, achieving an average performance of 66.4\%, 83.1\%, and 63.4\% AUROC on Real3D-AD, Anomaly-ShapeNet, and MulSen-AD, respectively. 

\end{abstract}

\section{Introduction}

\begin{figure}[t]
    \centering
    \includegraphics[width=\linewidth]{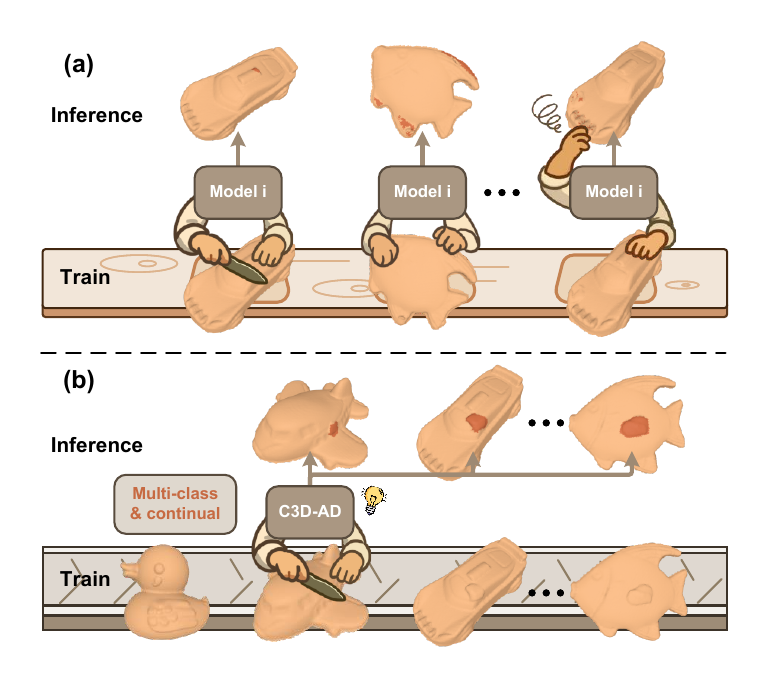}
    \caption{Difference between class-specific model and C3D-AD. (a) Single-class anomaly detection. (b) Multi-class and continual anomaly detection.}
    \label{fig: abstract}
\end{figure}

3D Anomaly Detection has garnered significant attention in identifying industrial product defects~\cite{tu2025lsfa, AST}. To detect anomalies effectively, feature-embedding methods and reconstruction-based methods have been proposed in recent years~\cite{2d1}. These methods extract features that are highly relevant for anomaly detection~\cite{BTF, tu2025lsfa, Liang2025ISMP}, leverage complementary modality information to enhance performance~\cite{AST, Gu2024MMRD}, reconstruct point cloud to allow greater generalization across data distribution~\cite{easynet, zhou2025r3dad}, and enhance the capability of multi-class anomaly detection~\cite{wang2025exploiting, cheng2025mc3dadunifiedgeometryawarereconstruction}. The performance of anomaly detection methods has steadily improved, and the application scenarios have become increasingly close to reality. 

However, real-world environments present a more complex challenge, e.g., the continuous emergence of new object categories requires detection. In this context, class-specific models are inefficient as they necessitate complete retraining for each new class. Unified models, while more efficient, are susceptible to catastrophic forgetting, causing the degradation of performance on previously learned tasks. Hence, strategies to mitigate this problem are crucial, which leads to the paradigm of Continual Learning (CL). CL aims to develop models able to learn sequentially from new data without forgetting previously acquired knowledge, as shown in Figure~\ref{fig: abstract}.

Continual learning has been applied to 2D anomaly detection. The rehearsal-based methods, e.g., CAD~\cite{Li2022CAD} and ReplayCAD~\cite{hu2025replaycad}, store the information of the Gaussian distribution and parameter of the diffusion model, respectively. The regularization-based methods constrain the parameters to be less sensitive to the new training data. For example, CDAD~\cite{Li2025one4more} constrains the gradients orthogonal to previous feature representations. Tang et al.~\cite{tang2025unifiedad} used a semantic compression strategy to maximize the space margin from different tasks. However, due to high-resolution inputs and class-specific models, the methods based on continual learning cannot be directly applied to 3D point clouds. 

Motivated by the observations above, we propose Continual 3D Anomaly Detection (C3DAD), a novel framework designed to address the challenges of sequential anomaly detection in point cloud data. Specifically, we introduce a Kernel Attention with random feature Layer (KAL) to extract generalized features. Rather than directly encoding raw point-level information, KAL extracts the spatial context of the point cloud and mines local structure in the unified kernel space. To efficiently preserve and update multi-class data cache across tasks and reconstruct data correctly and continually, we further propose a Kernel Attention with learnable Advisor (KAA) mechanism for the Encoder-Decoder module. Hence, the module can learn the information from new categories and discard redundant information. Moreover, to mitigate catastrophic forgetting and keep the representation consistent, we proposed the Reconstruction with Parameter Perturbation (RPP) module. This module encourages the model returns category-adaptive representations across all sequential tasks. The main contributions are as follows:
\begin{itemize}
    \item We introduce a novel layer named KAL to normalize the feature space while extracting features. Leveraging KAL, features are extracted from the unified kernel space, which is significant to continual 3D AD. By extracting local structure information, our method significantly enhances the model's ability for continual learning.
    \item We proposed a novel network KAA for continual learning. To address the limitations of fixed-capacity networks in continual learning, we propose a novel encoder-decoder architecture with learnable advisors that reduces redundant information from previous knowledge while learning new knowledge. Traditional attention mechanisms often encounter $\mathcal{O}(n^2)$ complexity, leading to computational bottlenecks. To overcome it, KAA with linear $\mathcal{O}(n)$ complexity updates advisors without compromising effectiveness.
    \item We present a new hypothesis constraint in continual learning. The network in its current state should retain satisfactory performance on past data when revisited. To enforce this, the network’s future outputs are predicted and constrained between current outputs by RPP, preserving the model’s continual learning abilities over time.
\end{itemize}

\section{Related Work}

\subsection{3D Anomaly Detection}
3D anomaly detection is a computer vision task focused on detecting and scoring anomalous points within 3D data, such as point clouds~\cite{Liu2023real3d, li2023scalable3danomalydetection, Li2025mulsen}, to identify product defects in industrial manufacturing. This is achieved through two main methods.

\textbf{(1) Feature-embedding methods}~\cite{BTF, Kruse2024SplatPose, Liang2025ISMP} extract embeddings from 3D data and measure similarity to normal data. Specifically, Student-Teacher networks~\cite{AST, Bergmann20233dst, QIN2023tsfew, Gu2024MMRD} effectively assess output differences to indicate anomaly levels. In contrast, \textbf{(2) reconstruction-based methods} evaluate errors by comparing outputs to original inputs~\cite{easynet, liang2025dunet}, allowing greater generalization across data distributions. Researchers have explored various reconstruction frameworks; for example, Masuda et al.~\cite{masuda2021vae3d} proposed an unsupervised anomaly detection framework based on VAE~\cite{kingma2022vae}, and Chen et al.~\cite{easynet} introduced a novel encoder-decoder for multi-scale and multimodal data. R3D-AD~\cite{zhou2025r3dad}, based on diffusion~\cite{ho2020diffusion}, obscures anomalous geometry for global anomaly detection. Recently, the unified model MC3D-AD~\cite{cheng2025mc3dadunifiedgeometryawarereconstruction} demonstrated significant performance improvements for multi-class data, highlighting the value of a one-for-all approach.

However, existing methods struggle to generalize in class-increasing settings. To address this, we propose a novel continual learning framework (C3D-AD) that enables multi-class and continual anomaly detection.

\subsection{Continual Learning}

Continual learning (CL) is a machine learning paradigm that enables models to learn continuously from evolving data streams, adapting to dynamic scenarios without full retraining. It provides lifelong learning capabilities while addressing catastrophic forgetting through three main methods~\cite{Mallya_2018_ECCV}. \textbf{(1) Regularization-based methods} constrain parameter updates to preserve crucial knowledge~\cite{Rebuffi_2017_CVPR}. For instance, Elastic Weight Consolidation uses the Fisher Information Matrix to identify critical parameters, while Learning without Forgetting employs knowledge distillation~\cite{batra2024evclelasticvariationalcontinual}. The STAR method constrains gradient updates using buffered samples~\cite{eskandar2025star}. \textbf{(2) Rehearsal-based methods} store and reuse past samples, effectively mitigating catastrophic forgetting~\cite{10.1609/aaai.v33i01.33011352}. LiDER enhances network smoothness by optimizing Lipschitz constants~\cite{bonicelli2022LiDER}. \textbf{(3) Architecture-based methods} allocate specific parameters for each task~\cite{mallya2018packnetaddingmultipletasks}. The LPS algorithm by Wang et al. partitions the network into task-specific sections to retain information from new tasks~\cite{wang2020LPS}.

CL has been successfully applied to 2D anomaly detection~\cite{barusco2025memory}. For example, Li et al.~\cite{Li2025one4more} proposed CDAD, which projects gradients into a subspace orthogonal to previous feature representations, while Tang et al.~\cite{tang2025unifiedad} used a semantic compression strategy to retain essential memories. CAD~\cite{Li2022CAD} and ReplayCAD~\cite{hu2025replaycad} utilize rehearsals of statistical information from previous distributions. Additionally, Liu et al. introduced a continual prompt module in UCAD~\cite{Liu2024ucad} for task adaptation.

However, due to high-resolution inputs and class-specific models, these methods cannot be directly applied to 3D data, necessitating retraining when encountering new categories. The development of a unified model for 3D anomaly detection in a class-incremental manner has yet to be explored. Therefore, we propose C3D-AD to facilitate the reconstruction of multi-class data and the continual detection of anomalies.

\section{Method}

\begin{figure*}
    \centering
    \includegraphics[width=\linewidth]{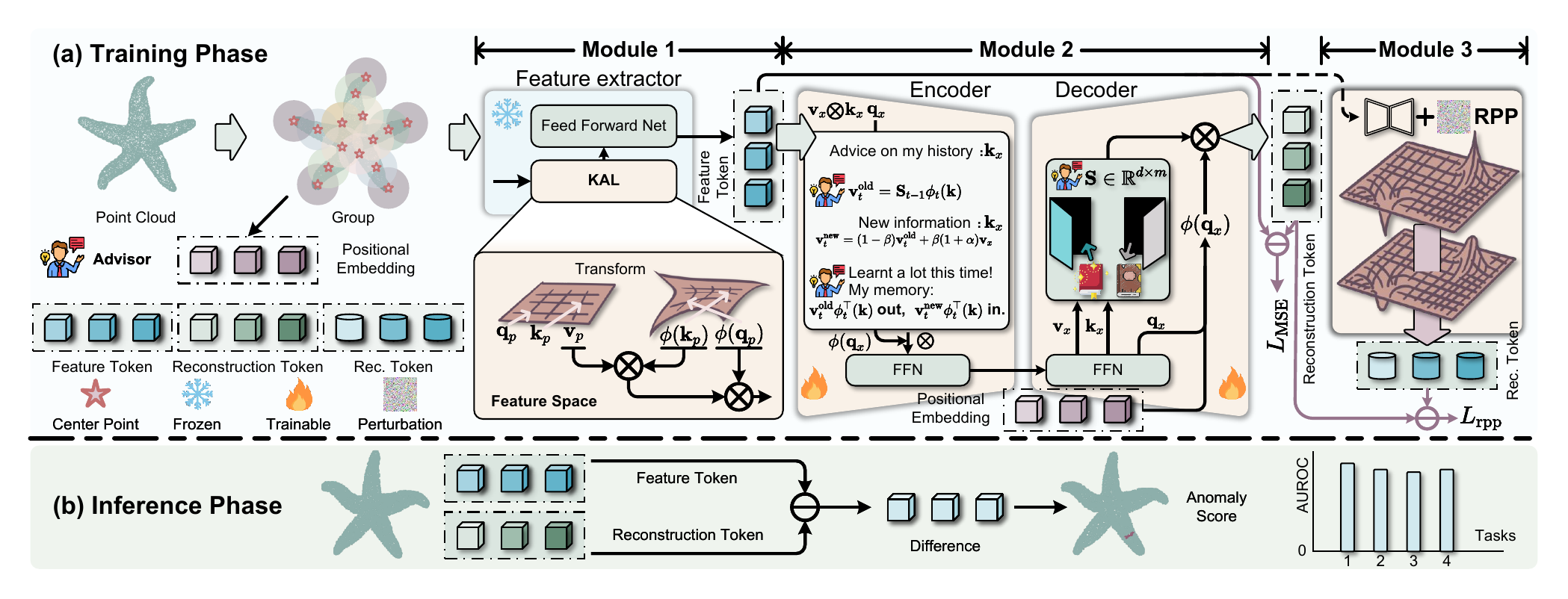}
    \caption{\textbf{The pipeline of C3D-AD.} The training point cloud data is aggregated into groups according to the center points. Feature tokens are generated by extracting features from both group centers and point groups, utilizing the Kernel Attention with random feature Layer (KAL) module in linear complexity. Then, feature tokens are input into the Encoder-Decoder, employing Kernel Attention with learnable Advisor (KAA) mechanism, which can memorize new class information and discard redundant information. Finally, the feature tokens are reconstructed again via Reconstruction with Parameter Perturbation (RPP) module, which can help the model review past samples and avoid catastrophic forgetting. Anomalies are detected according to the anomaly score by comparing the differences between feature tokens and reconstruction tokens.}
    \label{fig: pipeline}
\end{figure*}

\subsection{Problem Statement}
The task of continual 3D AD is to find a model to detect data from new categories while avoiding catastrophic forgetting. Considering the real application, the model can access only the training data of the current task and memory bank, precluding the storage or revisiting of full data from previous tasks.
Hence, the available data arrives sequentially in multiple tasks, i.e., $\mathcal{P}_{\textrm{train}} = \mathcal{P}^1_{\textrm{train}} \cup \mathcal{P}^2_{\textrm{train}} \cup \cdots \cup \mathcal{P}^T_{\textrm{train}} $ and $\mathcal{P}^i_{\textrm{train}} \cap \mathcal{P}^j_{\textrm{train}} = \emptyset~(i \ne j)$, where $\mathcal{P}^t_{\textrm{train}}$ denotes the training data from the $t$-th task containing point cloud samples from specific categories and only having normal samples, and $T$ is the total number of tasks. In the testing phase, the data to be detected includes both normal and anomalous point cloud samples from all encountered tasks, i.e., $ \mathcal{P}^1_{\textrm{test}} \subset \mathcal{P}^2_{\textrm{test}} \subset \cdots \subset \mathcal{P}^T_{\textrm{test}} = \mathcal{P}_{\textrm{test}}$. Our objective is to train a unified model that can detect anomalies across all encountered tasks while avoiding catastrophic forgetting of previously learned detection capabilities, which minimizes the cost of training the model. 

\subsection{Preliminary}
In this paper, vectors, matrices, and sets are denoted by the bold lowercase letters (e.g., $\mathbf{p}$), bold uppercase letters (e.g., $\mathbf{P}$), and calligraphic fonts (e.g., $\mathcal{P}$). 
Given the point cloud set $\mathcal{P}$, each point is represented as $\mathbf{p}_i \in \mathbb{R}^3, (i = 1, \dots, N)$. The point cloud is organized into a feature matrix $\mathbf{P} \in \mathbb{R}^{N \times 3}$. Points cloud from $\mathcal{P}$ can be encoded into tokens, yielding a token feature matrix $\mathbf{X} \in \mathbb{R}^{n \times d}$ with $n$ tokens. $\mathbf{W} \in \mathbb{R}^{d \times r}$ is a linear projection matrix.
 
\subsection{Overview Framework}
The key challenge in continual 3D anomaly detection is constraining models to avoid catastrophic forgetting of past detection capabilities when facing new data. To this end, we propose a novel \textbf{C}ontinual \textbf{3D} \textbf{A}nomaly \textbf{D}etection~(C3DAD). The overall framework is illustrated in Figure~\ref{fig: pipeline}, comprising three main components: 
a Kernel Attention with random feature Layer (KAL), 
a Kernel Attention with learnable Advisor (KAA), and Reconstruction with Parameter Perturbation (RPP).

\subsection{Kernel Attention with random feature Layer}
Feature extraction is used to improve the performance of the downstream tasks. In the continual learning paradigm, it is essential to extract generalized features. These features aim to minimize the feature space differences across different tasks, mitigating catastrophic forgetting. In the scenario of 3D AD, point clouds suffer from poor structure and weak semantic information due to the organization as triplets in the form of $\mathbf{P} \in \mathbb{R}^{N\times3}$. Hence, firstly, the local structure features are extracted for the generalized representation of the point clouds.

Specifically, $\mathbf{P}$ is sampled into $n$ centers group $\bar{\mathbf{P}}_{\textrm{center}}$ by the Furthest Point Sampling (FPS)~\cite{fps}. After sampling center points from the point cloud, the neighborhood point set of center point $\bar{\mathbf{p}}_i$ can be expressed as:
\begin{equation}
    \mathcal{N}_r(\bar{\mathbf{p}}_i) = \{ \mathbf{p}_j \in \mathbf{P} ~|~ \| \bar{\mathbf{p}}_i - \mathbf{p}_j \|_2 \le r \}.
\end{equation}
Due to varying point cloud scales across different classes, the radius $r$ is adaptively adjusted using the following equation:
\begin{equation}
    r = \frac{\eta}{|\bar{\mathbf{P}}_{\textrm{center}}|} \sum_{\bar{\mathbf{p}_j} \in \bar{\mathbf{P}}_{\textrm{center}}} \| \bar{\mathbf{p}}_i - \bar{\mathbf{p}}_j \|_2,
\end{equation}
where $\eta$ is the scaling factor of radius. 

Secondly, the features should be mapped into a unified space. 
To tokenize the point cloud data, an encoder is employed. Without loss of generality, let $\mathbf{q}$, $\mathbf{k}$, and $\mathbf{v}$ denote the query, key, and value vectors, respectively. The traditional self-attention mechanism utilized in encoders is:
\begin{equation}
    \mathbf{o}_l = \sum_{i}^{n} \frac{\exp(\mathbf{q}_l^{\top} \mathbf{k}_i)}{\sum_{j}^{n} \exp(\mathbf{q}_l^{\top} \mathbf{k}_j)} \mathbf{v}_i.
    \label{eq: tradi_attn}
\end{equation}
Without considering the \textit{softmax} activation function and scaling, the output is the optimum of $\min_{\mathbf{O}}\| \mathbf{O} - \mathbf{Q} \mathbf{K}^{\top} \mathbf{V} \ \|_{F}$, where $\mathbf{K}$ and $\mathbf{V}$ can be viewed as feature extraction and reconstruction matrix. 
The following kernel attention is proposed to capture nonlinear relationships among features:
\begin{equation}
    \mathbf{o}_l = \sum_{i}^{n} \frac{\kappa(\mathbf{q}_l, \mathbf{k}_i)}{\sum_{j}^{n} \kappa(\mathbf{q}_l, \mathbf{k}_j)} \mathbf{v}_i,
\end{equation}
where $\kappa(\mathbf{Q}, \mathbf{K})_{lj}=\kappa(\mathbf{q}_l, \mathbf{k}_j)$ is the kernel function. $\kappa(\mathbf{q}_l, \mathbf{k}_j)$ represents the inner product $\left \langle \phi(\mathbf{q}_l), \phi(\mathbf{k}_j) \right \rangle $, where $\phi(\cdot)$ is a mapping to the unified Hilbert space. If the space spanned by $\phi(\mathbf{K})$ is approximately unified, it can extract global generalized features in the continual learning paradigm. The output of the attention layer is:
\begin{equation}
    \mathbf{o}_l = \sum_{i}^{n} \frac{\phi^{\top}(\mathbf{q}_l) \phi(\mathbf{k}_i)}{\sum_{j}^{n} \phi^{\top}(\mathbf{q}_l) \phi(\mathbf{k}_i)} \mathbf{v}_i.
\end{equation}
However, the overall computational complexity is quadratic. To address this, it can be:
\begin{equation}
    \mathbf{o}_l = \frac{\left(\sum_{i}^{n} \mathbf{v}_i \phi^{\top}(\mathbf{k}_i)\right) \phi(\mathbf{q}_l)}{\sum_{j}^{n} \phi^{\top}(\mathbf{k}_j) \phi(\mathbf{q}_l)},
    \label{eq: linear_attn}
\end{equation}
which is $\mathcal{O}(n)$ computational complexity in matrix form. The mapping $\phi$ can be defined by the random feature, e.g., Positive Random Feature~\cite{choromanski2021rethinking}:
\begin{equation}
    \phi(\mathbf{x})=\frac{e^{\frac{-\|\mathbf{x}\|_2^2}{2}} \left[e^{\mathbf{w}_1^{\top} \mathbf{x}}, \cdots, e^{\mathbf{w}_m^{\top} \mathbf{x}}\right]}{\sqrt{m}}, 
\end{equation}
where the projection $\mathbf{w}_i$ is sampled i.i.d. from $\mathbf{w}_i \sim \mathcal{N}\left(0, I_d\right)$. Usually, $m$ is set to a small enough value. 

In this way, KAL not only captures the local spatial context of the point cloud but also learn the generalized nonlinear structural information across inter-group point clouds.

\subsection{Kernel Attention with learnable Advisor}
After extracting generalized representations of point clouds locally and globally via the KAL, it is essential to introduce an advisor within the encoder-decoder architecture to mitigate catastrophic forgetting during continual learning. 
To address this, we introduce a novel continual learning attention mechanism with linear $\mathcal{O}(n)$ complexity, named Kernel Attention with learnable Advisor (KAA), which enables the model to learn new information efficiently while preserving previously acquired information.

Rewrite Eq. \eqref{eq: linear_attn} by ignoring the denominator, and it becomes $\mathbf{o}_l = \mathbf{S} \phi(\mathbf{q}_l)$. $\mathbf{S} \in \mathbb{R}^{d \times m}$ is a continually learnable advisor. The following objective function \eqref{eq: obj_func} is proposed to train the advisor $\mathbf{S}$:
\begin{equation}
    \min_{\mathbf{S}} L_{\textrm{kaa}}(\mathbf{S}) = \min_{\mathbf{S}} \frac{1}{2} \|\mathbf{S}\phi(\mathbf{k}) - \mathbf{v}\|^2 - \alpha \operatorname{Tr}(\mathbf{v}^{\top}\mathbf{S}\phi(\mathbf{k})),
    \label{eq: obj_func}
\end{equation}
where the advisor $\mathbf{S}$ guides the key $\phi(\mathbf{k})$ close to the value $\mathbf{v}$ and aligns their directions.
Getting the derivative of the function $L_{\textrm{kaa}}$ w.r.t. $\mathbf{S}$ and setting it to zero, the update gradient can be derived as $\nabla_{\mathbf{S}} L_{\textrm{kaa}} = \mathbf{S}\phi(\mathbf k)\phi^{\top}(\mathbf{k}) - (1+\alpha)\phi(\mathbf k)\mathbf{v}^{\top}$. Given the learning rate $\beta$, the update rule is: 
\begin{equation}
    \begin{aligned}
        \mathbf{S}_t & = \mathbf{S}_{t-1} - \beta \nabla_{\mathbf{S}} L_{\textrm{kaa}} \\
        & = \mathbf{S}_{t-1} - \beta (\mathbf{S}_{t-1}\phi(\mathbf{k})\phi^{\top}(\mathbf{k}) - (1+\alpha)\phi(\mathbf{k})\mathbf{v}^{\top}).
    \end{aligned}
    \label{eq: update_rule}
\end{equation}
Let $\mathbf{v}_t^{\mathrm{old}} = \mathbf{S}_{t-1} \phi_t(\mathbf{k})$ and $\mathbf{v}_t^{\mathrm{new}} = (1-\beta)\mathbf{v}_t^{\mathrm{old}} + \beta (1+\alpha)\mathbf{v}_t$, then Eq. \eqref{eq: update_rule} becomes:
\begin{equation}
    \mathbf{S}_t = \mathbf{S}_{t-1} - \mathbf{v}_t^{\mathrm{old}}\phi_t^{\top}(\mathbf{k}) + \mathbf{v}_t^{\mathrm{new}}\phi_t^{\top}(\mathbf{k}),
    \label{eq: continual_update_rule}
\end{equation}
where $-\mathbf{v}_t^{\mathrm{old}}\phi_t^{\top}(\mathbf{k}) $ represents that reduces redundant information from previous tasks and $\mathbf{v}_t^{\mathrm{new}}\phi_t^{\top}(\mathbf{k})$ is to learn new information.
Hence, the output of attention in the $t$-th task is:
\begin{equation}
    \mathbf{O}_t = \phi(\mathbf{Q}_t) \mathbf{S}_t^{\top},
\end{equation}
where $l$-th row of $\phi(\mathbf{Q}_t)$ is $\phi^{\top}(\mathbf{q}_l)$. 

Storing historical point cloud information in $\mathbf{S}$ preserves KAA's continual learning capability. It eliminates the need to maintain a set of past samples. Furthermore, due to its linear complexity, the proposed method is resource-efficient.

\subsection{Reconstruction with Parameter Perturbation}
KAA is to learn new information while reducing previous one for the Encoder-Decoder. In this section, the Reconstruction with Parameter Perturbation (RPP) mechanism is proposed to reconstruct data from the view of the future.
To ensure that the model converges globally to a hypothesis space that is optimal across all tasks, the gradient of the model’s parameters must be continually constrained to optimize within the intersection of the optimal hypotheses for each task. This strategy aims to prevent catastrophic forgetting. According to the objective, the model should maintain a high similarity for the same batch of data at time $t$ and time $t + \Delta t$, $\Delta t > 0$. Consequently, this is formulated as the following minimization of a loss:
\begin{equation}
    L_{\textrm{rpp}} = \| h(\theta_t, x) - h(\theta_{t+\Delta t}, x) \|_2^2.
\end{equation}
where $h(\theta_t, \cdot)$ is the hypothesis at time $t$.
However, it is impossible to obtain the hypothesis $h(\theta_{t+\Delta t}, \cdot)$ at time $t$. Hence, the state after $\Delta t$ must be predicted based on the current model weights. To this end, we approximate the hypothesis after $\Delta t$ as $h(\theta_t + \delta)$, where the initial perturbation $\delta$ is sampled from a normal distribution and constrained by $\|\delta\|_2 \le \epsilon$. Within the parameter space, the perturbation is searched to induce the worst-case deviation upon addition, thereby simulating the most adverse future scenario. Motivated by this, the optimization problem is proposed:
\begin{equation}
\begin{aligned}
    L_{\textrm{rpp}}(\theta_t, \mathbf{P}_t) & = \max_{\delta} \| h(\theta_t, \mathbf{P}_t) - h(\theta_t + \delta, \mathbf{P}_t) \|_2^2 \\ 
    & \textrm{s.t.}~ \|\delta\|_2 \le \epsilon.
\end{aligned}
\end{equation}
To find a local optimum of this objective function, gradient ascent is employed. 

\subsubsection{Generalization Error Bound of RPP Loss} To assess the generalization error incurred by the objective function when using limited point cloud data to converge the model to the concept set, the generalization bound is derived for $L_{\textrm{rpp}}$.
\begin{theorem}
    Let $\mathcal{G} = \{g: x \mapsto \|h(\theta, x) - h(\theta + \delta, x)\|_2^2 \mid \|\delta\|_2 \le \epsilon \}$ be the class of function induced by bounded perturbations $\delta$. For all $g \in \mathcal{G}$ and any input $x$, with probability at least $1 - \xi$ over the random draw of a training sample of size $N$ from the underlying distribution $\mathcal{D}$, the following holds:
    \begin{equation}
        \hat{L}_{\textrm{rpp}} (\theta) \le L_{\textrm{rpp}}(\theta) + 2\epsilon^2 L_{\theta}^2 \mathscr{R}(\psi) + 3\sqrt{\frac{\log (2 / \xi)}{2 N}}.
    \end{equation}
    \label{theor: genbound}
\end{theorem}
\noindent According to Theorem~\ref{theor: genbound}, the perturbation constraint $\epsilon$ should not be excessively large to avoid a significant increase in the generalization error. More details can be found in the \textit{Supplemental Materials}. 

\section{Experiments}

\begin{table*}[htp]
\small
\centering
\begin{tabular}{lccccccccccc}
\toprule
& \multicolumn{4}{c}{\textit{Real3D-AD}} & \multicolumn{4}{c}{\textit{Anomaly-ShapeNet}} & \multicolumn{3}{c}{\textit{MulSen-AD}} \\
\midrule
task\_id & 1 & 2 & 3 & 4 & 1 & 2 & 3 & 4 & 1 & 2 & 3 \\ 
\midrule
Continual-Reg3D-AD & 0.598 & 0.527 & 0.505 & 0.521 & 0.522 & 0.544 & 0.501 & 0.519 & 0.602 & 0.599 & 0.580 \\
Continual-PatchCore (FPFH) & 0.566 & 0.527 & 0.500 & 0.518 & 0.567 & 0.541 & 0.510 & 0.494 & 0.491 & 0.549& 0.524\\
Continual-MC3D-AD & 0.767 & 0.650 & 0.629 & 0.625 & 0.829 & 0.803 & 0.833 & 0.797 & 0.623 & 0.626 & 0.604 \\
\textbf{C3D-AD} & \textbf{0.805} & \textbf{0.658} & \textbf{0.666} & \textbf{0.664} & \textbf{0.854} & \textbf{0.848} & \textbf{0.862} & \textbf{0.831} & \textbf{0.698} & \textbf{0.650} & \textbf{0.634} \\
\bottomrule
\end{tabular}
\caption{The mean O-AUROC ($\uparrow$) performance of different methods across multiple tasks of datasets. The best results are \textbf{bold}.}
\label{tab: continual_result}
\end{table*}

\begin{table*}[!ht]
\centering
\scriptsize
\begin{tabular}{@{}lcccccccccccccc@{}}
\toprule
\textbf{Method} & \textbf{cap0} & \textbf{cap3} & \textbf{helmet3} & \textbf{cup0} & \textbf{bowl4} & \textbf{vase3} & \textbf{headset1} & \textbf{eraser0} & \textbf{vase8} & \textbf{cap4} & \textbf{vase2} & \textbf{vase4} & \textbf{helmet0} & \textbf{bucket1} \\
\midrule
BTF(Raw) & 0.668 & 0.527 & 0.526 & 0.403 & 0.664 & 0.717 & 0.515 & 0.525 & 0.424 & 0.468 & 0.410 & 0.425 & 0.553 & 0.321 \\
BTF(FPFH) & 0.618 & 0.522 & 0.444 & 0.586 & 0.609 & 0.699 & 0.490 & 0.719 & 0.668 & 0.520 & 0.546 & 0.510 & 0.571 & 0.633 \\
M3DM & 0.557 & 0.423 & 0.374 & 0.539 & 0.464 & 0.439 & 0.617 & 0.627 & 0.663 & 0.777 & 0.737 & 0.476 & 0.526 & 0.501 \\
Patchcore(FPFH) & 0.580 & 0.453 & 0.404 & 0.600 & 0.494 & 0.449 & 0.637 & 0.657 & 0.662 & 0.757 & 0.721 & 0.506 & 0.546 & 0.551 \\
Patchcore(PointMAE) & 0.589 & 0.476 & 0.424 & 0.610 & 0.501 & 0.460 & 0.627 & 0.677 & 0.663 & 0.727 & 0.741 & 0.516 & 0.556 & 0.561 \\
CPMF & 0.601 & 0.551 & 0.520 & 0.497 & 0.683 & 0.582 & 0.458 & 0.689 & 0.529 & 0.553 & 0.582 & 0.514 & 0.555 & 0.601 \\
Reg3D-AD & 0.693 & 0.725 & 0.367 & 0.510 & 0.663 & 0.650 & 0.610 & 0.343 & 0.620 & 0.643 & 0.605 & 0.500 & 0.600 & 0.752 \\
IMRNet & 0.737 & 0.775 & 0.573 & 0.643 & 0.676 & 0.700 & 0.676 & 0.548 & 0.630 & 0.652 & 0.614 & 0.524 & 0.597 & 0.771 \\
R3D-AD & \textit{0.822} & 0.730 & 0.707 & 0.776 & 0.744 & 0.742 & \textit{0.795} & \textit{0.890} & 0.721 & 0.681 & 0.752 & 0.630 & \textit{0.757} & 0.756 \\
MC3D-AD & 0.793 & 0.701 & \textbf{0.979} & 0.743 & \underline{0.911} & 0.761 & \underline{0.886} & 0.776 & 0.670 & \underline{0.835} & \textit{0.929} & \textbf{0.876} & 0.672 & 0.784 \\
PLANE & \textbf{0.944} & \textbf{0.954} & 0.721 & \textit{0.805} & \textit{0.893} & \textit{0.782} & 0.776 & \textbf{1.000} & \textbf{0.964} & 0.730 & \textbf{0.971} & \textit{0.773} & 0.704 & \textbf{0.968} \\
PO3AD & \underline{0.877} & \underline{0.859} & \textit{0.754} & \underline{0.871} & \textbf{0.981} & \textbf{0.821} & \textbf{0.923} & \underline{0.995} & \textit{0.739} & \textit{0.792} & \underline{0.952} & 0.675 & \underline{0.762} & \textit{0.787} \\
\textbf{C3D-AD} & 0.678  & \textit{0.800} & \underline{0.900} & \textbf{0.929} & 0.626 & \underline{0.791} & {0.752} & {0.843} & \underline{0.773} & \textbf{0.884} & {0.919} & \underline{0.836} & \textbf{0.870} & \underline{0.863} \\
\bottomrule
\end{tabular}

\begin{tabular}{@{}lcccccccccccccc@{}}
\toprule
\textbf{Method} & \textbf{bottle3} & \textbf{vase0} & \textbf{bottle0} & \textbf{tap1} & \textbf{bowl0} & \textbf{bucket0} & \textbf{vase5} & \textbf{vase1} & \textbf{vase9} & \textbf{ashtray0} & \textbf{bottle1} & \textbf{tap0} & \textbf{phone} & \textbf{cup1} \\
\midrule
BTF(Raw) & 0.568 & 0.531 & 0.597 & 0.573 & 0.564 & 0.617 & 0.585 & 0.549 & 0.564 & 0.578 & 0.510 & 0.525 & 0.563 & 0.521 \\
BTF(FPFH) & 0.322 & 0.342 & 0.344 & 0.546 & 0.509 & 0.401 & 0.409 & 0.219 & 0.268 & 0.420 & 0.546 & 0.560 & 0.671 & 0.610 \\
M3DM & 0.541 & 0.423 & 0.574 & 0.739 & 0.634 & 0.309 & 0.317 & 0.427 & 0.663 & 0.577 & 0.637 & \textit{0.754} & 0.357 & 0.556 \\
Patchcore(FPFH) & 0.572 & 0.455 & 0.604 & 0.766 & 0.504 & 0.469 & 0.417 & 0.423 & 0.660 & 0.587 & 0.667 & 0.753 & 0.388 & 0.586 \\
Patchcore(PointMAE) & 0.650 & 0.447 & 0.513 & 0.538 & 0.523 & 0.593 & 0.579 & 0.552 & 0.629 & 0.591 & 0.601 & 0.458 & 0.488 & 0.556 \\
CPMF & 0.405 & 0.451 & 0.520 & 0.697 & 0.783 & 0.482 & 0.618 & 0.345 & 0.609 & 0.353 & 0.482 & 0.359 & 0.509 & 0.499 \\
Reg3D-AD & 0.525 & 0.533 & 0.486 & 0.641 & 0.671 & 0.610 & 0.520 & 0.702 & 0.594 & 0.597 & 0.695 & 0.676 & 0.414 & 0.538 \\
IMRNet & 0.640 & 0.533 & 0.552 & 0.696 & 0.681 & 0.580 & 0.676 & \textit{0.757} & 0.594 & 0.671 & 0.700 & 0.676 & 0.755 & 0.757 \\
R3D-AD & {0.781} & {0.788} & {0.733} & \underline{0.900} & {0.819} & {0.683} & {0.757} & 0.729 & {0.718} & {0.833} & {0.737} & 0.736 & {0.762} & {0.757} \\
MC3D-AD & {0.756} & {0.821} & {0.795} & \textbf{0.970} & \underline{0.930} & \underline{0.898} & \textbf{0.976} & \underline{0.857} & \textit{0.736} & \underline{0.962} & {0.709} & \underline{0.945} & \textit{0.919} & \underline{0.952} \\
PLANE & \textbf{0.994} & \textbf{0.896} & \textit{0.843} & 0.652 & \textbf{0.963} & \textbf{0.981} & 0.690 & \textit{0.771} & 0.592 & \textit{0.905} & \underline{0.814} & 0.467 & \textbf{1.000} & 0.705 \\
PO3AD & \underline{0.926} & \textit{0.858} & \underline{0.900} & 0.681 & \textit{0.922} & \textit{0.853} & \underline{0.852} & 0.742 & \underline{0.830} & \textbf{1.000} & \textbf{0.933} & 0.745 & 0.776 & \textit{0.833} \\
\textbf{C3D-AD} & \textit{0.857} & \underline{0.888} & \textbf{0.962} & \textit{0.844} & {0.919} & {0.822} & \textit{0.795} & \textbf{0.938} & \textbf{0.897} & {0.752} & \textit{0.740} & \textbf{0.955} & \underline{0.938} & \textbf{0.967} \\
\bottomrule
\end{tabular}

\begin{tabular}{@{}lcccccccccccc|cc@{}}
\toprule
\textbf{Method} & \textbf{vase7} & \textbf{helmet2} & \textbf{cap5} & \textbf{shelf0} & \textbf{bowl5} & \textbf{bowl3} & \textbf{helmet1} & \textbf{bowl1} & \textbf{headset0} & \textbf{bag0} & \textbf{bowl2} & \textbf{jar} & \textbf{Mean} & \textbf{A. R.} \\
\midrule
BTF(Raw) & 0.448 & 0.602 & 0.373 & 0.164 & 0.417 & 0.385 & 0.349 & 0.264 & 0.378 & 0.410 & 0.525 & 0.420 & 0.493 & 10.68\\
BTF(FPFH) & 0.518 & 0.542 & 0.586 & 0.609 & 0.699 & 0.490 & 0.719 & 0.668 & 0.520 & 0.546 & 0.510 & 0.424 & 0.528 & 9.98\\
M3DM & 0.657 & 0.623 & 0.639 & 0.564 & 0.409 & 0.617 & 0.427 & 0.663 & 0.577 & 0.537 & 0.684 & 0.441 & 0.552 & 9.73\\
Patchcore(FPFH) & 0.693 & 0.425 & \textit{0.790} & 0.494 & 0.558 & 0.537 & 0.484 & 0.639 & 0.583 & 0.571 & 0.615 & 0.472 & 0.568 & 9.25\\
Patchcore(PointMAE) & 0.650 & 0.447 & 0.538 & 0.523 & 0.593 & 0.579 & 0.552 & 0.629 & 0.591 & 0.601 & 0.458 & 0.483 & 0.562 & 9.35\\
CPMF & 0.397 & 0.462 & 0.697 & 0.685 & 0.685 & 0.658 & 0.589 & 0.639 & 0.643 & 0.643 & 0.625 & 0.610 & 0.559 & 9.25\\
Reg3D-AD & 0.462 & 0.614 & 0.467 & 0.688 & 0.593 & 0.348 & 0.381 & 0.525 & 0.537 & 0.706 & 0.490 & 0.592 & 0.572 & 9.30\\
IMRNet & 0.635 & {0.641} & 0.652 & 0.603 & {0.710} & 0.599 & 0.600 & 0.702 & 0.720 & 0.660 & 0.685 & 0.780 & 0.661 & 6.73\\
R3D-AD & {0.771} & {0.633} & 0.670 & {0.696} & 0.656 & {0.767} & {0.720} & {0.778} & {0.738} & {0.720} & {0.741} & {0.838} & {0.749} & {4.63}\\
MC3D-AD & \underline{0.938} & 0.609 & {0.761} & \textbf{0.841} & {0.754} & \textbf{0.885} & \textbf{1.000} & \textbf{0.978} & \textbf{0.862} & {0.805} & {0.719} & \underline{0.971} & \underline{0.842} & \textit{3.05}\\
PLANE & \underline{0.938} & \textbf{1.000} & \textbf{0.870} & \textit{0.759} & \textit{0.800} & 0.706 & 0.543 & \underline{0.907} & 0.782 & \textbf{0.914} & \textbf{0.956} & \textbf{1.000} & 0.836 & 3.18 \\
PO3AD & \textbf{0.966} & \underline{0.869} & 0.670 & 0.573 & \underline{0.849} & \underline{0.881} & \textit{0.961} & \textit{0.829} & \textit{0.808} & \textit{0.833} & \underline{0.833} & 0.866 & \textit{0.839} & \underline{2.88} \\
\textbf{C3D-AD} & {0.900} & \textit{0.745} & \underline{0.811} & \underline{0.765} & \textbf{0.944} & \textit{0.793} & \underline{0.971} & {0.711} & \underline{0.827} & \underline{0.905} & \textit{0.822} & \textit{0.924} & \textbf{0.846} & \textbf{2.78}\\
\bottomrule
\end{tabular}
\caption{The O-AUROC ($\uparrow$) performance of different methods across 40 categories of Anomaly-ShapeNet. The best, second-best, and third-best are \textbf{bold}, \underline{underline}, and \textit{italics}, respectively. \textbf{A. R.} represents the average ranking of each method.}
\label{tab:auroc_results_no_color}
\end{table*}


\subsection{Experimental setting}
\subsubsection{Datasets.} The \textbf{Real3D-AD}~\cite{Liu2023real3d} is a benchmark for 3D AD, comprising 1,254 large-scale, high-resolution samples from 12 object categories. The training set for each category consists of only 4 normal samples, and the test set contains both normal samples and various anomalies. The \textbf{Anomaly-ShapeNet}~\cite{li2023scalable3danomalydetection} is a large-scale synthetic dataset for AD. It comprises 1,600 samples distributed across 40 object categories, posing a significant challenge due to its high inter-class diversity. \textbf{MulSen-AD}~\cite{Li2025mulsen} is the high-resolution multisensor anomaly detection data set. It consists of 2,035 samples from 15 industrial object categories, which are split into a training set of 1,391 normal samples and a test set comprising 150 normal and 494 anomalous samples. 

\subsubsection{Comparing baselines.} We selected classical 3D anomaly detection methods: BTF~\cite{BTF}, M3DM~\cite{M3DM}, Patchcore~\cite{Patchcore}, CPMF~\cite{CPMF}, Reg3D-AD~\cite{Liu2023real3d}, IMRNet~\cite{li2023scalable3danomalydetection}, R3D-AD~\cite{zhou2025r3dad}, MC3D-AD~\cite{cheng2025mc3dadunifiedgeometryawarereconstruction}, PLANE~\cite{wang2025exploiting}, and PO3AD~\cite{PO3AD} to demonstrate that C3D-AD is effective in anomaly detection. Moreover, we modified three baseline methods adapted for continual learning for fair comparison: Continual-PatchCore, Continual-MC3D-AD, and Continual-Reg3D-AD. Complete implementation details are provided in the \textit{Supplementary Materials}.

\subsubsection{Continual Learning Setting}
To estimate the models' performance in class-incremental 3D Anomaly Detection, we partitioned each dataset into a series of sequential tasks. For the Real3D-AD dataset, we divide the training set into 4 disjoint tasks based on categories. The test sets are constructed cumulatively, i.e., $\mathcal{P}_{\textrm{test}}^1 \subset \cdots \subset \mathcal{P}_{\textrm{test}}^4$. Following this, Anomaly-ShapeNet and MulSen-AD datasets are divided into 4 and 3 tasks, respectively. To evaluate the performance of the model, we adopt metrics at the object level. For object-level AD, the AUROC ($\uparrow$) is employed. To ensure a fair comparison across all categories, the mean AUROC and average ranking of each method ($\downarrow$) are reported. Code would be available upon acceptance of the paper.

\subsubsection{Implementation Details.} PointMAE~\cite{pointmae} with the KAL is pre-trained on ModelNet408K~\cite{Wu2015modelnet} for feature extraction. The $\alpha$ and $\beta$ of KAA are both set to $0.7$, and $m$ is set to $10$. The sample scaling factor $\eta$ is $10$ to cover the whole point cloud. The $\epsilon$ is scanned in the range $[0.01, 10]$, determining the generalization error. The AdamW optimizer is employed in the training process with initial learning rate $0.0001$ and rate $0.00001$ after 800 epochs. The number of stacked encoder-decoder blocks is set to $4$. Our experiments were conducted on a machine with PyTorch 1.13.0, CUDA 11.7, and an NVIDIA A100-PCIE-40GB GPU.



\subsection{Performance on Continual Anomaly Detection}
The experimental results of C3D-AD and other methods in the continual learning paradigm are shown in Table~\ref{tab: continual_result}. C3D-AD has demonstrated state-of-the-art (SOTA) performance in 3D Anomaly Detection across various tasks and datasets. Unlike memory-bank-based methods, e.g., continual-Reg3D-AD and continual-PatchCore, C3D-AD achieves superiority by unifying the feature space via KAL, while minimizing reconstruction error via RPP. Hence, C3D-AD outperform these two method by 14.3\%, 31.2\%, and 5.4\% on three dataset. Furthermore, while the unified model MC3D-AD implicitly maintains previous information, C3D-AD performs better. The gaps, which are 3.9\%, 3.4\%, and 3.0\%, are attributed to the lack of mechanisms like KAA encoding past information into the model's parameters and the RPP module for information rehearsal. The experiments also demonstrate that C3D-AD can be applied in class-incremental AD.

\begin{figure}[h]
    \centering
    \includegraphics[width=0.8\linewidth]{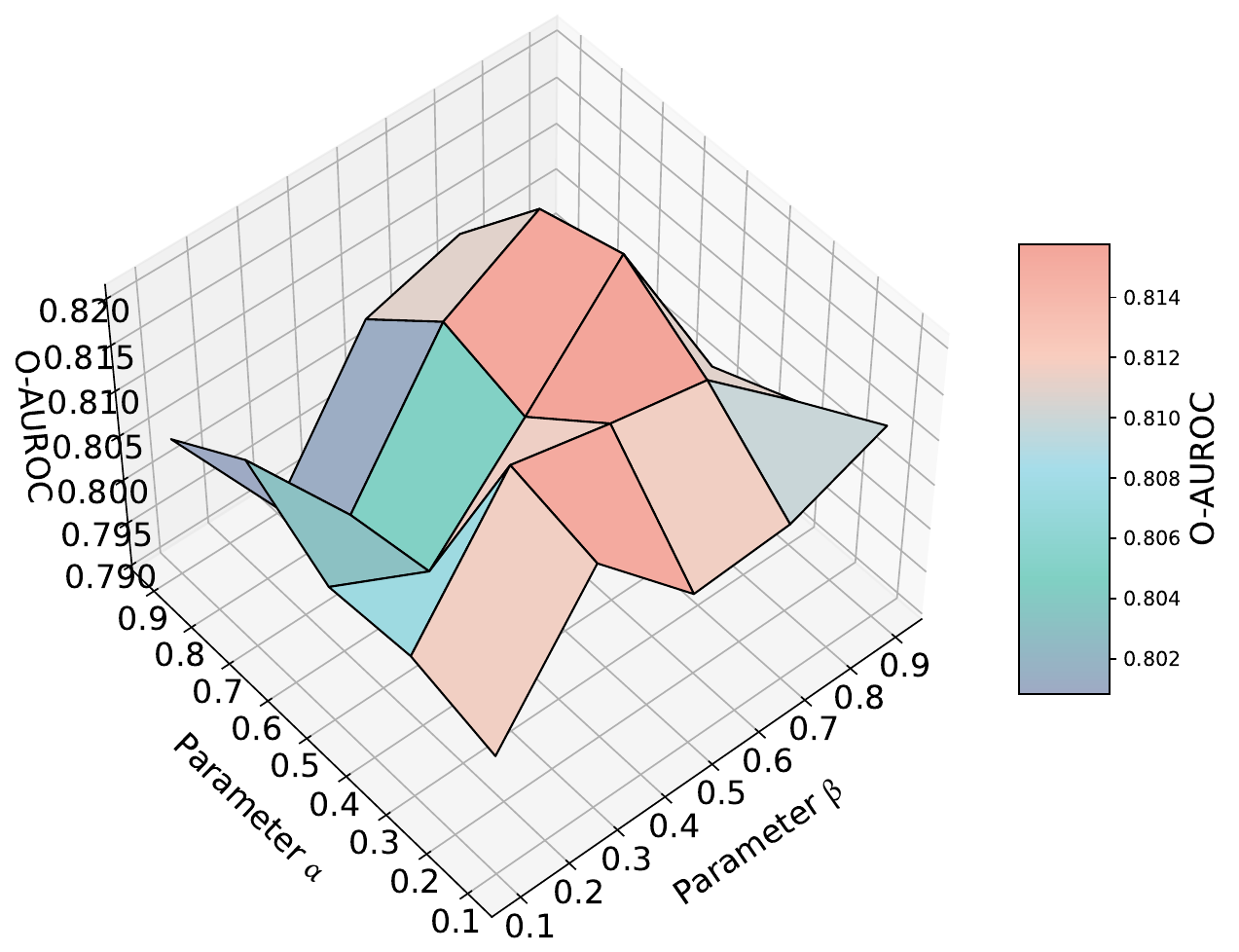}
    \caption{Sensitivity of $\alpha$ and $\beta$ on Anomaly-ShapeNet.}
    \label{fig: a_n_b}
\end{figure}

\subsection{Performance on Multi-class Anomaly Detection}
The capability of C3D-AD for continual 3D AD indicates that it has the intrinsic ability to learn from various data classes. To access the baseline of C3D-AD, multi-class anomaly detection is conducted, which is not on a continual learning paradigm. The experimental results are shown in Table~\ref{tab:auroc_results_no_color}. Compared to other methods, C3D-AD has the superiority on average AUROC, improving $0.4\%$ than MC3D-AD. In addition, C3D-AD achieves the top average ranking in 40 classes. This demonstrates its satisfactory performance in multi-class anomaly detection scenarios.

\subsection{Ablation Study}
To demonstrate the effectiveness of our proposed modules, ablation studies are conducted on KAL, KAA, and RPP. The experimental results, as presented in Table~\ref{tab: abla_real}, show that the model achieves its best performance on continual 3D AD only when all three components are integrated. Based on the results across the datasets, KAL and RPP show a more significant performance. This suggests that extracting generalized features and maintaining feature consistency are key to enhancing the continual 3D AD. Furthermore, the information from the advisor is also significant for the model to learn new data and avoid catastrophic forgetting continually.  

\begin{table}[h]
\centering
\begin{tabular}{ccc|ccc}
\toprule
KAL & KAA & RPP & R. & A. & M. \\
\midrule
$\times$ & $\checkmark$ & $\checkmark$ & 0.534 & 0.791 & 0.612 \\
$\checkmark$ & $\times$ & $\checkmark$ & 0.565 & 0.814 & 0.607 \\
$\checkmark$ & $\checkmark$ & $\times$ & 0.529 & 0.798 & 0.557 \\
\midrule
$\checkmark$ & $\checkmark$ & $\checkmark$ & \textbf{0.664} & \textbf{0.831} & \textbf{0.634} \\
\bottomrule 
\end{tabular}
\caption{Results of ablation study on Real3D-AD (R.), Anomaly-ShapeNet (A.), and MulSen-AD (M.).} 
\label{tab: abla_real}
\end{table}

\begin{table}[!ht]
\small
\centering
\begin{tabular}{lcccccc}
\toprule
$\epsilon=$ & 10$^{-2}$ & 10$^{-1}$ & 10$^{0}$ & 10$^{1}$ & 10$^{2}$ & 10$^{3}$ \\
\midrule
AUROC & 0.801 & 0.817 & 0.787 & 0.809 & 0.780 & 0.769 \\
\bottomrule 
\end{tabular}
\caption{Sensitivity analysis of $\epsilon$ on Anomaly-ShapeNet.} 
\label{tab: epsilon}
\end{table}

\begin{table}[!ht]
\small
\centering
\begin{tabular}{lcccc}
\toprule
$m=$ & 10$^{1}$ & 10$^{2}$ & 10$^{3}$ & 5$\times$10$^{3}$ \\
\midrule
AUROC & 0.802 & 0.828 & 0.842 & 0.819 \\
Inf. T. (s) & 0.281 & 0.311 & 0.393 & 1.109 \\
Memory (GB) & 4.431 & 4.650 & 5.680 & 9.400 \\
\bottomrule 
\end{tabular}
\caption{Analysis of $m$ on Anomaly-ShapeNet. The inference time (Inf. T.) and the GPU Memory usage are reported. } 
\label{tab: m}
\end{table}

\subsection{Sensitivity of Parameters}
The sensitivity experiments evaluate the value of hyperparameters. In KAA, $\alpha$ is to align the direction of information from the advisor and the value vector, and $\beta$ is to leverage the previous and new information. As shown in Figure \ref{fig: a_n_b}, $\alpha$ and $\beta$ are set to $0.7$, which can achieve satisfactory performance. In addition, $\epsilon$ is an important hyperparameter of RPP. According to Theorem~\ref{theor: genbound}, $\epsilon$ should not be too large, which is demonstrated as shown in Table~\ref{tab: epsilon}. The hyperparameter $m$ in KAL and KAA is usually a small value. As shown in Table~\ref{tab: m}, the computational complexity w.r.t. $m$ is $\mathcal{O}(m)$. However, larger $m$, which is the attribute of the advisor, may not lead to better performance. To leverage the performance and efficiency, $m$ of KAL and KAA can be set from 10$^{1}$ to 10$^{3}$.

\begin{figure}[t!]
    \centering
    \includegraphics[width=0.8\linewidth]{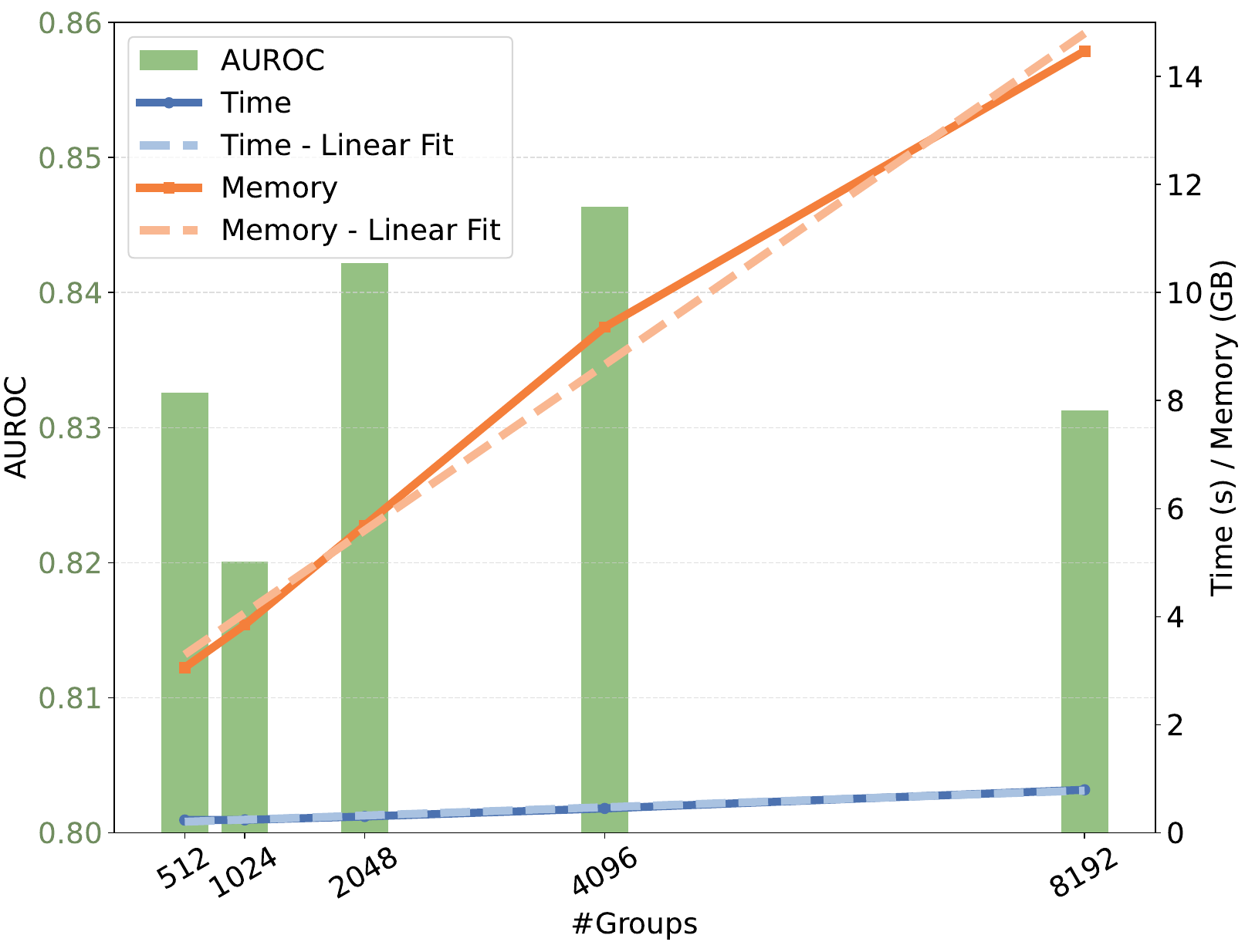}
    \caption{Inference time (s) and memory usage (GB) vs. \#Groups for C3D-AD on Anomaly-ShapeNet.}
    \label{fig: on}
\end{figure}

\subsection{Empirical Computational Complexity Analysis}
Due to the linear computational complexity of Kernel Attention in KAL and KAA, C3D-AD is efficient in anomaly detection. Figure~\ref{fig: on} shows that the inference time and GPU peak memory increase linearly with respect to the group number, which is relevant to the number of tokens. The performance can be satisfactory by setting the number to $4096$. Hence, C3D-AD is tailored for industrial continual 3D AD due to its $\mathcal{O}(n)$ complexity and performance.

\section{Conclusions}

The continuous emergence of new object categories poses a significant challenge for 3D Anomaly Detection. To address this, in this study, we propose a continual learning framework named C3D-AD. Firstly, we introduce a Kernel Attention with random feature Layer to extract generalized features. Then, to reconstruct feature tokens while avoiding catastrophic forgetting, a Kernel Attention with learnable Advisor module is designed within the encoder-decoder to learn new information while discarding redundant one. Furthermore, to maintain representation consistency across tasks, the tokens are reconstructed again using Reconstruction with Parameter Perturbation, which aligns the model's current and future outputs. Experiments on benchmark datasets demonstrate our method's superiority in continual 3D AD, achieving state-of-the-art performance with satisfactory efficiency. For future work, further research is needed to explore how to effectively constrain the advisor to achieve better performance in continual 3D AD.

\bibliography{aaai2026}


\end{document}